\documentclass[10pt,twocolumn,letterpaper]{article}

\usepackage{cvpr}
\usepackage{times}
\usepackage{epsfig}
\usepackage{graphicx}
\usepackage{amsmath}
\usepackage{amssymb}

\usepackage{multirow}
\usepackage{booktabs}
\usepackage{hhline}
\usepackage{multirow}
\usepackage{subfigure}
\usepackage{hyperref}

\cvprfinalcopy 

\def\cvprPaperID{1552} 

\ifcvprfinal\fi
\begin{document}

\title{Global versus Localized Generative Adversarial Nets}

\author{Guo-Jun Qi, Liheng Zhang, Hao Hu, Marzieh Edraki\\
{\tt\small guojun.qi@ucf.edu, \{lihengzhang1993,hao\_hu,m.edraki\}@knights.ucf.edu}\\
\and
Jingdong Wang, and Xian-Sheng Hua\\
{\tt\small welleast@outlook.com, xiansheng.hxs@alibaba-inc.com}\\
Laboratory for MAchine Perception and LEarning (MAPLE)\\
\url{maple.cs.ucf.edu}\\
University of Central Florida and Alibaba Group
}

\maketitle

\begin{abstract}
In this paper, we present to learn a novel localized Generative Adversarial Net (GAN) on the manifold of real data. Compared with the classic GAN that {\em globally} parameterizes a manifold, the Localized GAN (LGAN) uses local coordinate charts to parameterize local geometry of data transformations across different locations on the manifold. Specifically, around each point there exists a {\em local} generator to produce diverse data following various patterns of transformations along the manifold.  The locality nature of LGAN enables it to directly access the local geometry with no need to invert the generator in the classic GAN to access its global coordinates. Furthermore, it can prevent the manifold from being locally collapsed to be dimensionally deficient by imposing an orthonormality prior between tangents. This provides a geometric approach to alleviating mode collapse on the manifold at least locally by preventing vanishing or dependent data variations along different coordinates. We will also demonstrate the LGAN can be applied to train a locally consistent classifier that is robust against perturbations along the manifold, and the resultant regularizer is closely related to the Laplace-Beltrami operator without relying on an approximate graph-based manifold representation. Our experiments show that the proposed LGANs can not only produce diverse image transformations, but also deliver superior classification performances.
%
\end{abstract}

\section{Introduction}
The classic Generative Adversarial Net (GAN) \cite{goodfellow2014generative} seeks to generate samples with indistinguishable distributions from real data.  For this purpose, it learns a generator $G(\mathbf z)$ as a function that maps from input random noises $\mathbf z$ drawn from a distribution $P_\mathcal Z$ to output data $G(\mathbf z)$. A discriminator is learned to distinguish between real and generated samples. The generator and discriminator are jointly trained in an adversarial fashion so that the generator fools the discriminator by improving the quality of generated data.

All the samples produced by the learned generator form a manifold $\mathcal M=\{G(\mathbf z)|\mathbf z\sim P_\mathcal Z\}$, with the input variables $\mathbf z$ as its global coordinates. However, a global coordinate system could be too restrictive to capture various forms of local transformations on the manifold. For example, a nonrigid object like human body and a rigid object like a car admit different forms of variations on their shapes and appearances, resulting in distinct geometric structures unfit into a single coordinate chart of image transformations.

Indeed, existence of a global coordinate system is a too strong assumption
for many manifolds. 
For example, there does not exist a global coordinate chart covering an entire hyper-sphere embedded in a high dimensional space as it is even not topologically similar (i.e., homeomorphic) to an Euclidean space. This prohibits the existence of a global isomorphism between a single coordinate space and the hyper-sphere, making it impossible to study the underlying geometry in a global coordinate system.
For this reason, mathematicians instead use an atlas of {\em local} coordinate charts located at different points on a manifold to study the underlying geometry \cite{willmore2013introduction}. 

Even when a global coordinate chart exists, a global GAN could still suffer two serious challenges. First, a point $\mathbf x$ on manifold cannot be directly mapped back to its global coordinates $\mathbf z$, i.e., finding $\mathbf z$ such as $G(\mathbf z)=\mathbf x$ for a given $\mathbf x$.  But many applications need the coordinates of a given point $\mathbf x$ to access its local geometry such as tangents and curvatures. Thus, for a global GAN, one has to solve the inverse $G^{-1}$ of a generator network (e.g., via an autoencoder such as VAE \cite{kingma2013auto}, ALI \cite{dumoulin2016adversarially} and BiGAN \cite{donahue2016adversarial}) to access the coordinates of a point $\mathbf x$ and then its local geometry of data transformations along the manifold.





The other problem is the manifold generated by a global GAN could locally collapse. Geometrically, on a $N$-dimensional manifold, this occurs if the tangent space $\mathcal T_\mathbf x$ of a point $\mathbf x$  is dimensionally deficient, i.e., ${\rm dim}~\mathcal T_\mathbf x < N$ when tangents become linearly dependent along some coordinates \footnote{For example, on a $2$-D surface, the manifold reduces to an $1$-D curve or a $0$-D singularity at some points.}.
In this case, data variations become redundant or even vanish along some directions on the manifold.
Moreover, a locally collapsed tangent space at a point $\mathbf x$ could be related with a {\em collapsed mode} \cite{goodfellow2014generative,salimans2016improved}, around which a generator $G(\mathbf z)$ would no longer produce diverse data as $\mathbf z$ changes in different directions. This provides us with an alternative geometric insight into mode collapse phenomena observed in literature \cite{radford2015unsupervised}.



\begin{figure}[!t]
\vspace{2mm}
\begin{center}
\includegraphics[width=0.8\linewidth]{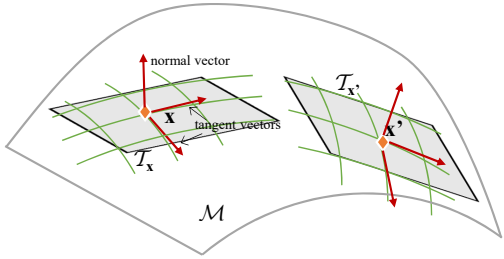}
\end{center}
 \caption{Illustration of a curved manifold $\mathcal M$ embedded in $3$-dimensional ambient space. At each location $\mathbf x$, its tangent space $\mathcal T_\mathbf x$ consists of all tangent vectors  to the manifold. These tangent vectors characterize the geometry of local transformations allowed to move a point $\mathbf x$ on $\mathcal M$.
 }\label{Fig:diff_local}
\end{figure}

The above challenges inspire us to develop a Localized GAN (LGAN) by learning local generators $G(\mathbf x, \mathbf z)$ associated with individual points $\mathbf x$ \footnote{At first glance, the form of a local generator $G(\mathbf x, \mathbf z)$ looks like a conditional GAN (cGAN) with $\mathbf x$ as its condition. However, a local generator in LGAN intrinsically differs from cGAN in its geometric representation of a local coordinate chart. Refer to Section~\ref{sec:lgan} for details.
} on a manifold. As illustrated in Figure~\ref{Fig:diff_local}, local generators are located around different data points so that the pieces of data generated by different local generators can be sewed together to cover an entire manifold seamlessly. Different pieces of generated data are not isolated but could have some overlaps between each other to form a connected manifold \cite{qi2011locality}.


The advantage of the LGAN is at least twofold. First, one can directly access the local geometry of transformations near a point without having to evaluate its global coordinates, as each point is directly localized by a local generator in the corresponding local coordinate chart.
This {\em locality} nature of LGAN makes it straightforward to explore pointwise geometric properties across a manifold.
Moreover, we will impose an orthonormality prior on the local tangents, and the resultant orthonormal basis spans a full dimensional tangent space, preventing a manifold from being locally collapsed.
It allows the model to explore diverse patterns of data transformations disentangled in different directions, leading to a geometric approach at least {\em locally} alleviating the mode collapse problem on a manifold.



We will also demonstrate an application of the LGAN to train a robust classifier by encouraging
a smooth change of the classification decision on the manifold formed by the LGAN.
The classifier is trained with a regularizer that minimizes the square norm of the classifier's gradient on the manifold, which is closely related with Laplace-Beltrami operator.
The local coordinate representation in LGAN makes it straightforward to train such a classifier with no need of computing global coordinates of training examples to access their local geometry of transformations.  Moreover, the learned orthonormal tangent basis also allows the model to effectively explore various forms of independent transformations allowed on the underlying manifold.



The remainder of this paper is organized as follows. In the next section, we will review the related works, followed by Section~\ref{sec:lgan} in which we present the proposed Localized GANs.  In Section~\ref{sec:lgan-ssl}, a semi-supervised learning algorithm is presented to use LGANs to train a robust classifier that is locally consistent over the manifold formed by a LGAN model.

\section{Related Works}
{\bf \noindent Global vs. Localized GANs.} By different types of coordinate systems used to parameterize their data manifolds,  we can categorize the GANs into {\em global} and {\em local} models. Existing models, including the seminal GAN model proposed by Goodfellow et al. \cite{goodfellow2014generative,qi2017loss} and many variants \cite{wgan17}, are {\em global} GANs that use a global coordinate chart to parameterize the generated data. In contrast, the localized GAN presented in this paper is a local paradigm, which uses local coordinate charts centered at different data points to form a manifold by a collection of local generators.

The distinction between global and local coordinate systems results in conceptual and algorithmic differences between global and local GANs. Conceptually, the global GANs assume that the manifolds formed by their generators could be globally parameterized in a way that the manifolds are topologically similar to an Euclidean coordinate space.  In contrast, the localized paradigm abandons the global parameterizability  assumption, allowing us to use multiple local coordinate charts to cover an entire manifold. Algorithmically, if a global GAN needs to access local geometric information underlying its generated manifold, it has to invert the generator function in order to find the global coordinates corresponding to a given data point.  This is usually performed by learning an auto-encoder network along with the GAN models, e.g., BiGAN \cite{donahue2016adversarial} and ALI \cite{dumoulin2016adversarially}. On the contrary, the localized GAN enables direct access to local geometry without having to invert a global generator, since the local geometry around a point is directly available from the corresponding local generator. Moreover, the orthormornality between local tangents could also maximize the capability of a local GAN in exploring independent local transformations along different coordinate directions, thereby preventing the manifold of generated data from being locally collapsed with deficient dimensions.

{\bf\noindent Semi-Supervised Learning.} One of the most important applications of GANs lies in the classification problem, especially considering their ability of modeling the manifold structures for both labeled and unlabeled examples \cite{tang2007typicality,tang2008integrated,qi2017joint}. For example, \cite{kingma2014semi} presented variational auto-encoders \cite{kingma2013auto} by combining deep generative models and approximate variational inference to explore both labeled and unlabeled data.
\cite{salimans2016improved} treated the samples from the GAN generator as a fake class, and explore unlabeled examples by assigning them to a real class different from the fake one. \cite{rasmus2015semi} proposed to train a ladder network \cite{valpola2015neural} by minimizing the sum of supervised and unsupervised cost functions through back-propagation, which avoids the conventional layer-wise pre-training approach. \cite{springenberg2015unsupervised} presented an approach to learning a discriminative classifier by trading-off mutual information between observed examples and their predicted classes against an adversarial generative model. \cite{dumoulin2016adversarially} sought to jointly distinguish between not only real and generated samples but also their latent variables in an adversarial fashion. 
\cite{dai2017good} presented to train a semi-supervised classifier by exploring the areas where real samples are unlikely to appear.

In this paper, we will explore the LGAN's ability of modeling the data distribution and its manifold geometry to train a robust classifier, which can make locally consistent classification decisions in presence of small perturbations on data. The idea of training a locally consistent classifier could trace back almost two decades ago to TangentProp \cite{simard2000transformation} that pursued classification invariance against image rotation and translation manually performed in an ad-hoc fashion.  Kumar et al. \cite{kumar2017improved} extended the TangentProp by training an augmented form of BiGAN to explore the underlying data distributions, but it still relied on a global GAN to indirectly access the local tangents by learning a separate encoder network.
On the contrary, the local coordinates will enable the LGAN to directly access the geometry of image transformations to train a {\em locally consistent} classifier, along with the orthonormality between local tangents allowing the learned classifier to explore its local consistency against independent image transformations along different local coordinates on the manifold.
\section{Localized GANs}\label{sec:lgan}
We present the proposed Localized GANs (LGANs). Before that, we first briefly review the classic GANs in the context of differentiable manifolds.
\subsection{Classic GAN and Global Coordinates}
A Generative Adversarial Net (GAN) seeks to train a generator $G(\mathbf z)$ by transforming a random noise $\mathbf z\in\mathbb R^N$ drawn from $P_\mathcal Z$ to a data sample $G(\mathbf z)\in\mathbb R^D$. Such a classic GAN uses a global $N$-dimensional coordinate system $\mathbf z$ to represent its generated samples $G(\mathbf z)$ residing in an ambient space $\mathbb R^D$.  Then all the generated samples form a $N$-dimensional manifold $\mathcal M=\{G(\mathbf z)|\mathbf z\in\mathbb R^N\}$ that is embedded in $\mathbb R^D$.

In a global coordinate system, the local structure (e.g., tangent vectors and space) of a given data point $\mathbf x$ is not directly accessible, since one has to compute its corresponding coordinates $\mathbf z$ to localize the point on the manifold.  One often has to resort to an inverse of the generator (e.g., via ALI and BiGAN) to find the mapping from $\mathbf x$ back to $\mathbf z$.

Even worse, the tangent space $\mathcal T_\mathbf x$ could locally collapse at a point $\mathbf x$ if it is dimensionally deficient (i.e., ${\rm dim}~\mathcal T_\mathbf x < N$). Actually, if ${\rm dim}~\mathcal T_\mathbf x$ is extremely low (i.e., $<<N$), a locally collapsed point $\mathbf x$ could become a collapsed mode on the manifold, around which $G(\mathbf z)$ would no longer produce significant data variations even though $\mathbf z$ changes in different directions. For example, if ${\rm dim}~\mathcal T_\mathbf x = 1$, there is only a curve of data variations passing through $\mathbf x$. In an extreme case ${\rm dim}~\mathcal T_\mathbf x = 0$, the data variations would completely vanish as $\mathbf x$ becomes a singular point on the manifold.


\subsection{Local Generators and Tangent Spaces}
Unlike the classic GAN, we propose a Localized GAN (LGAN) model equipped with a local generator $G(\mathbf x, \mathbf z)$ that can produce various examples in the neighborhood of a point $\mathbf x\in\mathbb R^D$ on the manifold.

This forms a local coordinate chart $\{G(\mathbf x, \mathbf z)|\mathbf z\subset\mathbb R^N\sim P_\mathcal Z\}$ around $\mathbf x$, with its local {\em coordinates} $\mathbf z$ drawn from a random distribution $P_\mathcal Z$ over an Euclidean space $\mathbb R^N$. In this manner, an atlas of local coordinate charts can cover an entire manifold $\mathcal M$ by a collection of local generators $G(\mathbf x,\mathbf z)$ located at different points on $\mathcal M$.

In particular, for $G(\mathbf x, \mathbf z)$, we assume that the origin of the local coordinates $\mathbf z$ should be located at the given point $\mathbf x$, i.e., $G(\mathbf x, \mathbf 0)=\mathbf x$, where $\mathbf 0\in\mathbb R^N$ is an all-zero vector.

To study the local geometry near a point $\mathbf x$, we need tangent vectors located at $\mathbf x$ on the manifold.
By changing the value of a coordinate $\mathbf z^j$ while fixing the others, the points generated by $G(\mathbf x, \mathbf z)$ form a coordinate curve passing through $\mathbf x$ on the manifold. Then, the vector tangent to this coordinate curve at $\mathbf x$ is
\begin{equation}\label{eq:tangent}
\boldsymbol\tau_\mathbf x^j \triangleq \frac{{\partial G(\mathbf x,\mathbf z)}}{{\partial {\mathbf z^j}}}{\mid_{\mathbf z = \mathbf 0}}\in\mathbb R^D.
\end{equation}

All such $N$ tangent vectors $\boldsymbol\tau_\mathbf x^j,j=1,\cdots,N$ form a basis spanning a linear tangent space $\mathcal T_\mathbf x={\rm Span}(\boldsymbol\tau_\mathbf x^1, \cdots,\boldsymbol\tau_\mathbf x^N)$ at $\mathbf x$.  This tangent space consists of all vectors tangent to some curves passing through $\mathbf x$ on the manifold. Each tangent $\boldsymbol \tau \in \mathcal T_\mathbf x$ characterizes some local transformation in the direction of this tangent vector.

A Jacobian matrix $\mathbf J_\mathbf x \in\mathbb R^{D\times N}$ can also be defined by stacking all $N$ tangent vectors $\boldsymbol\tau_\mathbf x^j$ in its columns.

\subsection{Regularity: Locality and Orthonormality}
However, there exists a challenge that the tangent space $\mathcal T_\mathbf x$ would collapse if it is dimensionally deficient, i.e, its dimension ${\rm dim} \mathcal T_\mathbf x$ is smaller than the manifold dimension $N$. If this occurs, the $N$ tangents in (\ref{eq:tangent}) could reduce to dependent transformations that would even vanish along some coordinates $\mathbf z$.

To prevent the collapse of the tangent space, we need to impose a regularity condition that the $N$ basis $\{\boldsymbol\tau_\mathbf x^j, j=1,\cdots,N\}$ of $\mathcal T_\mathbf x$ should be linearly independent of each other. This guarantees the manifold be locally ``similar" (diffeomorphic mathematically) to a $N$-dimensional Euclidean space, rather than being collapsed to a lower-dimensional subspace having dependent local coordinates.

As a linearly independent basis can always be transformed to an orthonormal counterpart by a proper transformation,
one can set the orthonormal condition on the tangent vectors $\boldsymbol\tau_\mathbf x^j$, i.e.,
\begin{equation}\label{eq:orth}
\langle\boldsymbol\tau_\mathbf x^i,\boldsymbol\tau_\mathbf x^j\rangle=\delta_{ij}
\end{equation}
where $\delta_{ij}=0$ for $i\neq j$ and $\delta_{ii}=1$ otherwise.  The resultant orthonormal basis of tangent vectors capture the independent components of local transformations near individual data points on the manifold.

In summary, the local generator $G(\mathbf x, \mathbf z)$ should satisfy the following two conditions:

\noindent { \em (i) locality:} $G(\mathbf x, \mathbf 0)=\mathbf x$, i.e., the origin of the local coordinates $\mathbf z$ should be located at $\mathbf x$;

\noindent { \em (ii) orthonormality:} $\mathbf J_\mathbf x^T \mathbf J_\mathbf x=\mathbf I_N$, which is a matrix form of (\ref{eq:orth}) with $I_N$ being the identity matrix of size $N$.

One can minimize the following regularizer on $G(\mathbf x, \mathbf z)$ to penalize the violation of these two conditions \footnote{Alternatively, we can parameterize $G(\mathbf x, \mathbf z)$ as $\mathbf x + B(\mathbf x, \mathbf z)-B(\mathbf x, \mathbf 0)$ with a network $B$ modeling a perturbation on $\mathbf x$. Such a parameterization of local generator directly satisfies the locality constraint $G(\mathbf x, \mathbf 0)=\mathbf x$. },
\begin{equation}\label{eq:reg}
\Omega_G(\mathbf x) = \mu \|G(\mathbf x, \mathbf 0)-\mathbf x\|^2 + \eta \|\mathbf J_\mathbf x^T\mathbf  J_\mathbf x - \mathbf I_N\|^2
\end{equation}
where $\mu$ and $\eta$ are nonnegative weighting coefficients for the two terms.
By using a deep network for computing $G(\mathbf x, \mathbf z)$, this regularizer can be minimized by backpropagation algorithm.


\subsection{Training $G(\mathbf x, \mathbf z)$}
Now the learning problem for the localized GANs boils down to train a $G(\mathbf x, \mathbf z)$. Like the GANs, we will train a discriminator $D(\mathbf x)$ to distinguish between real samples drawn from a data distribution $P_\mathcal X$ and generated samples by $G(\mathbf x, \mathbf z)$ with $\mathbf x\sim P_\mathcal X$ and $\mathbf z\sim P_\mathcal Z$ as follows.
\[
\mathop {\max }\limits_D {\mathbb E_{\mathbf x \sim P_\mathcal X}}\log D(\mathbf x) + {\mathbb E_{\mathbf x \sim P_\mathcal X,\mathbf z \sim P_\mathcal Z}}\log (1 - D(G(\mathbf x,\mathbf z))
\]
where $D(\mathbf x)$ is the probability of $\mathbf x$ being real, and the maximization is performed wrt the model parameters of discriminator $D$.

On the other hand, the generator can be trained by maximizing the likelihood that the generated samples by $G(\mathbf x, \mathbf z)$ are real as well as minimizing the regularization term (\ref{eq:reg}).

\[
\mathop {\min }\limits_G  - {\mathbb E_{\mathbf x \sim P_\mathcal X,\mathbf z \sim P_\mathcal Z}}\log D(G(\mathbf x,\mathbf z)) + {\mathbb E_{\mathbf x \sim P_\mathcal X}}{\Omega _G(\mathbf x)}
\]
where the minimization is performed wrt the model parameters of local generator $G$, and the regularization enforces the locality and orthonormality conditions on $G$.

Then $D$ and $G$ can be alternately optimized by stochastic gradient descent via a backpropagation algorithm.
\section{Semi-Supervised LGANs}\label{sec:lgan-ssl}
In this section, we will show that the LGAN can help us train a locally consistent classifier by exploring the manifold geometry. First we will discuss the functional gradient on a manifold in Section~\ref{sec:gradient}, and show its connection with Laplace-Beltrami operator that generalizes the graph Laplacian in Section~\ref{sec:laplace}. Finally, we will present the proposed LGAN-based classifier in detail in Section~\ref{sec:lganssl}.

\subsection{Functional Gradient along Manifold}\label{sec:gradient}

First let us discuss how to calculate the derive of a function on the manifold.

Consider a function $f(\mathbf x)$ defined on the manifold. At a given point $\mathbf x$, its neighborhood on the manifold is depicted by $G(\mathbf x, \mathbf z)$ with the local coordinates $\mathbf z$. By viewing $f$ as a function of $\mathbf z$, we can compute the derivative of $f$ when it is restricted on the manifold.

It is not hard to obtain the derivative of $f(G(\mathbf x, \mathbf z))$ with respect to a coordinate $\mathbf z^j$ by the chain rule,
$$
\dfrac{\partial f(G(\mathbf x, \mathbf z))}{\partial \mathbf z^j}|_{\mathbf z=\mathbf 0}
= \langle \boldsymbol \tau^j_\mathbf x, \nabla_\mathbf x f(\mathbf x)\rangle
$$
where
$\nabla_\mathbf x f(\mathbf x)$
is the gradient of $f$ at $\mathbf x$, and $\langle \cdot,\cdot \rangle$ is the inner product between two vectors.
It depicts how fast $f$ changes as a point moves away from $\mathbf x$ along the coordinate $\mathbf z^j$ on the manifold.

Then, the gradient of $f$ at $\mathbf x$ when $f$ is restricted on the manifold $G(\mathbf x, \mathbf z)$ can be written as
\begin{equation}\label{eq:change}
\begin{aligned}
\nabla^G_\mathbf x f\triangleq \nabla_\mathbf z f(G(\mathbf x,\mathbf z)) |_{\mathbf z=\mathbf 0}
= \mathbf J_\mathbf x^T \nabla_\mathbf x f(\mathbf x)
\end{aligned}
\end{equation}
Geometrically, it shows the gradient of $f$ along the manifold can be obtained by projecting the regular gradient $\nabla_\mathbf x f$ onto the tangent space $\mathcal T_\mathbf x$ with the Jacobian matrix $\mathbf J_\mathbf x$. Here we denote the resultant gradient {\em along manifold} by $\nabla^G_\mathbf x f$ to highlight its dependency on $G(\mathbf x, \mathbf z)$




\subsection{Connection with Laplace-Beltrami Operator}\label{sec:laplace}
If $f$ is a classifier, $\nabla_\mathbf z f(G(\mathbf x, \mathbf z))$ depicts the change of the classification decision on the manifold formed by $G(\mathbf x, \mathbf z)$. At $\mathbf x$, the change of $f$ restricted on $G(\mathbf x, \mathbf z)$ can be written as
\begin{equation}\label{eq:fchange}
\begin{aligned}
|f(G(\mathbf x, \mathbf z+\delta \mathbf z))-f(G(\mathbf x,\mathbf z))|^2 \approx \|\nabla^G_\mathbf x f \|^2\delta\mathbf z
\end{aligned}
\end{equation}
It shows that penalizing $\|\nabla^G_\mathbf x f \|^2$
can train a robust classifier that is resilient against a small perturbation $\delta\mathbf z$ on a manifold. It is supposed to deliver locally consistent classification results in presence of noises.

The functional gradient is closely related with the Laplace-Beltrami operator, the one that is widely used as a regularizer on the graph-based semi-supervised
learning \cite{belkin2004semi,belkin2006manifold,zhu2003semi,zhou2004learning}.

It is well known that the divergence operator $-\rm{div}$ and the gradient $\nabla$ are formally adjoint, i.e.,
$\int_\mathcal M \langle \mathbf V, \nabla^G_\mathbf x f\rangle d P_\mathcal X= \int_\mathcal M {\rm div}(\mathbf V)f d P_\mathcal X$. Thus we have
\begin{equation}\label{eq:laplace}
\int_\mathcal M \|\nabla^G_\mathbf x f\|^2 d P_\mathcal X
= \int_\mathcal M f {\rm div} (\nabla^G_\mathbf x f) d P_\mathcal X
\end{equation}
where $\Delta f\triangleq {\rm div} (\nabla^G_\mathbf x f) $  is the Laplace-Beltrami operator.

In graph-based semi-supervised learning, one constructs a graph representation of data points to approximate the underlying data manifold \cite{belkin2006manifold}, and then use
a Laplacian matrix to approximate the Laplace-Beltrami operator $\Delta f$.

In contrast, with the help of LGAN, we can directly obtain $\Delta f$ on $G(\mathbf x, \mathbf z)$ without having to
resort to a graph representation. Actually, as the tangent space at a point $\mathbf x$ has an orthonormal basis, we can write
\begin{equation}\label{eq:second}
\Delta f = {\rm div}(\nabla^G_\mathbf x f)=\sum_{j=1}^N \dfrac{\partial^2 f(G(\mathbf x, \mathbf z))}{\partial (\mathbf z^i)^2}
\end{equation}

In the following, we will learn  a locally consistent classifier on the manifold by penalizing a sudden change of its classification function $f$ in the neighborhood of a point. We can implement it by minimizing either the square norm of the gradient or the related Laplace-Beltrami operator. For simplicity, we will choose to penalize the gradient of the classifier as it only involves computing the first-order derivatives of a function compared with the Laplace-Beltrami operator having the higher-order derivatives.

\subsection{Locally Consistent Semi-Supervised Classifier}\label{sec:lganssl}
We consider a semi-supervised learning problem with a set of training examples $(\mathbf x_l, y_l)$ drawn from a distribution $P_\mathcal L$ of labeled data. We also have some unlabeled examples $\mathbf x_u$ drawn from the data distribution $P_\mathcal X$ of real samples.
The amount of unlabeled examples is often much larger  than their labeled counterparts, and thus can provide useful information for training $G$ to capture the manifold structure of real data.

Suppose that there are $K$ classes, and we attempt to train a classifier $P(y|\mathbf x)$ for  $y\in\{1,2,\cdots,K+1\}$ that outputs the probability of $\mathbf x$ being assigned to a class $y$ \cite{salimans2016improved}. The first $K$ are real classes and the last one is a fake class denoting $\mathbf x$ is a generated example.

This probabilistic classifier can be trained by the following objective function
\begin{equation}\label{eq:prob}
\begin{aligned}
 \mathop {\max }\limits_P ~&{\mathbb E_{({\mathbf x_l},{y_l}) \sim P_\mathcal L}}\log P({y_l}|{\mathbf x_l}) + {\mathbb E_{\mathbf x_u \sim P_\mathcal X}}\log P({y_u} \le K|{\mathbf x_u}) \\
  +& {\mathbb E_{\mathbf x \sim P_\mathcal X,\mathbf z \sim P_\mathcal Z}}\log P(y = K + 1|G(\mathbf x,\mathbf z)) \\
  -& \sum_{k=1}^K \mathbb E_{\mathbf x\sim P_\mathcal X} \|\nabla^G_\mathbf x \log P(y=k|\mathbf x)\|^2
 \end{aligned}
 \end{equation}
 where $\nabla^G_\mathbf x \log P(y=k|\mathbf x)$ of the last term is the gradient of the log-likelihood  along the manifold $G(\mathbf x, \mathbf z)$ at $\mathbf x$, that is
 $
 \nabla_\mathbf z \log P(y=k|G(\mathbf x, \mathbf z))|_{\mathbf z=\mathbf 0}.
 $
 Let us explain the objective (\ref{eq:prob}) in detail below.
\begin{itemize}
\item The first term maximizes the log-likelihood that a labeled training example drawn from the distribution $P_\mathcal L$ of labeled examples is correctly classified by $P(y|\mathbf x)$.
\item The second term maximizes the log-likelihood that an unlabeled example $\mathbf x_u$ drawn from the data distribution $P_\mathcal X$ is assigned to one of $K$ real classes (i.e., $y_u\leq K$).
\item The third term enforces $P(y|\mathbf x)$ to classify a generated sample by $G(\mathbf x, \mathbf z)$ as fake (i.e., $y=K+1$).
\item The last term penalizes a sudden change of classification function on the manifold, thus yielding a locally consistent classifier as expected. This can be seen by viewing $\log P(y|\mathbf x)$ as  $f$ in (\ref{eq:fchange}).
\end{itemize}

On the other hand, with a fixed classifier $P(y|\mathbf x)$, the local generator $G$ is trained by the following objective:
\begin{equation}\label{eq:gen}
\begin{aligned}
 \mathop {\min }\limits_G  ~ \mathcal K_G + \mathcal L_G + {\mathbb E_{\mathbf x \sim P_\mathcal X}}{\Omega _G(\mathbf x)}
 \end{aligned}
\end{equation}
where
\begin{itemize}
\item The first term is {\em label preservation} term
$$
\mathcal K_G=- {\mathbb E_{({\mathbf x_l},{y_l}) \sim P_\mathcal L,\mathbf z \sim P_\mathcal Z}}\log P({y_l}|G({\mathbf x_l},\mathbf z))
$$
which enforces generated samples should not change the labels of their original examples. This label preservation term can help explore intra-class variance by generating new variants of training examples without changing their labels.

\item The second term is feature matching loss
$\mathcal L_G=\|\mathbb E_{\mathbf x \sim P_\mathcal X} \boldsymbol\psi_P(\mathbf x)-\mathbb E_{\mathbf x \sim P_\mathcal X,\mathbf z \sim P_\mathcal Z} \boldsymbol\psi_P(G(\mathbf x, \mathbf z))\|^2$, where $\boldsymbol\psi_P$ is an intermediate layer of feature representation from the classification network $P$. It minimizes the feature discrepancy between real and generated examples, and exhibits competitive performance in literature \cite{salimans2016improved,kumar2017improved} for semi-supervised learning.

\item The third term is the regularizer $\Omega _G(\mathbf x)$ that enforces the locality and orthonormality priors on the local generator as shown in (\ref{eq:reg}).

\end{itemize}

\section{Experiments}
In this section, we conduct experiments to test the capability of the proposed LGAN on both image generation and classification tasks.

\begin{figure*}[htpb]
\centering
\subfigure{
\begin{minipage}{1.0\linewidth}
\begin{center}
	   \includegraphics[width=1.0\linewidth]{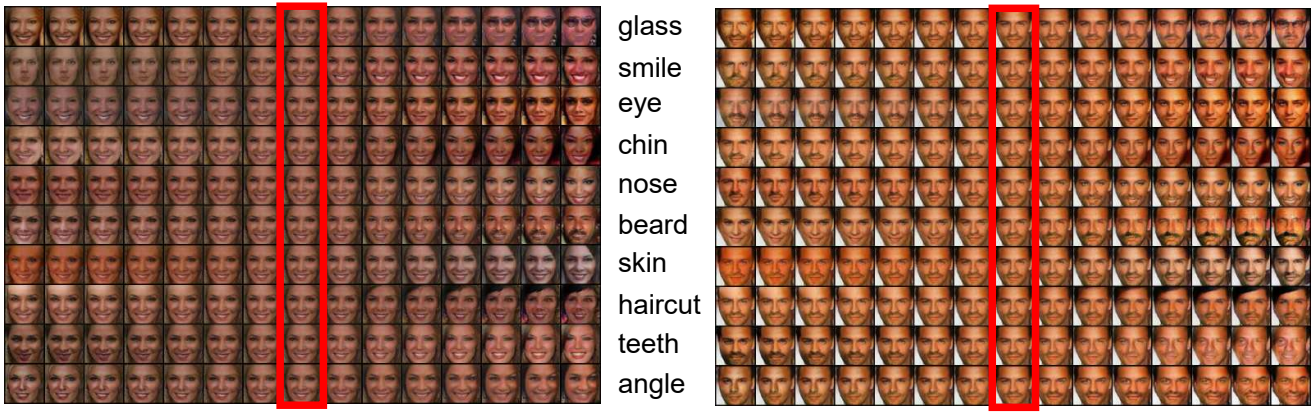}
\end{center}
\end{minipage}}
   \caption{Faces generated by LGAN on the CelebA dataset. The middle column in a red bounding box represents the image at the origin $\mathbf z=0$ of a local coordinate chart. In each row, the images are generated along a local coordinate. There exist various patterns of image variations across different rows of faces, including whether wearing glasses and the variations in expressions, eyes, haircuts and so forth.}\label{fig:celebA}
\label{Fig:dRNN}
\end{figure*}

\subsection{Architecture and Training Details}
In this section, we discuss the network architecture and training details for the proposed LGAN model in image generation and claudication tasks.

\begin{figure}[t]
\centering
\subfigure{
\begin{minipage}{1.0\linewidth}
\begin{center}
	   \includegraphics[width=1.0\linewidth]{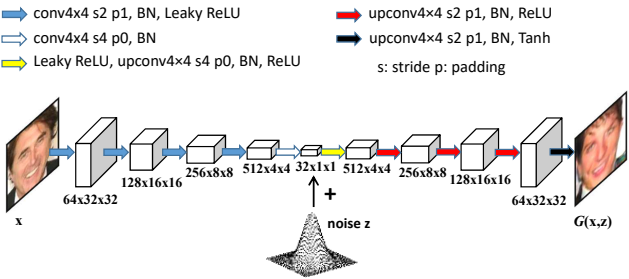}
\end{center}
\end{minipage}}
   \caption{Network architecture for local generators on the CelebA dataset.}\label{Fig:LG_arch}
\end{figure}

In experiments, the local generator network $G(\mathbf x, \mathbf z)$ was constructed by first using a CNN to map the input image $\mathbf x$ to a feature vector added with a noise vector of the same dimension. Then a deconvolutional network with fractional strides was used to generate output image $G(\mathbf x, \mathbf z)$. Figure \ref{Fig:LG_arch} illustrates the architecture for the local generator network used to produce images on CelebA. The same discriminator network as in DCGAN \cite{radford2015unsupervised} was used in LGAN.  The detail of network architectures used in semi-supervised classification tasks will be discussed shortly.

Instead of drawing $\mathbf z$ from a Gaussian distribution, the quality of generated images can be improved by training the LGAN with noises sampled from a mixture of Gaussian noise with a discrete distribution $\delta_\mathbf 0$ concentrated at $\mathbf 0$, i.e., $\mathbf z \sim 0.9~\mathcal N(\mathbf 0, \mathbf I) + 0.1~\delta_\mathbf 0$ where $\mathcal N(\mathbf 0, \mathbf I)$ is zero-mean Gaussian distribution with an identity covariance matrix $\mathbf I$. In other words, with a probability of $0.1$, $\mathbf z$ is set to $\mathbf 0$; otherwise, with probability of 0.9, it is drawn from $\mathcal N(\mathbf 0, \mathbf I)$. Sampling from $\delta_\mathbf 0$ could better serve to enforce the {\em locality} prior when training a local generator in its local coordinate chart.



We used Adam solver to update the network parameters where the learning rate is set to $5\times 10^{-5}$ and $10^{-3}$ for training discriminator and generator networks respectively. The two hyperparameters $\mu$ and $\eta$ imposing locality and orthonormality priors in the regularizer were chosen based on an independent validation set held out from the training set.

\subsection{Image Generation with Diversity}

Figure~\ref{fig:celebA} illustrates the generated images on the CelebA dataset.   In this task, $32$-D local coordinates were used in the LGAN, and each row was generated by varying one of $32$ local coordinates while fixing the others. In other worlds, each row represents image transformations in one coordinate direction.  The middle column in a red bounding box corresponds to the original image at the origin of local coordinates. The figure shows how a face transforms as it moves away from the origin along different coordinate directions on the manifold. The results demonstrate LGAN can generate sharp-looking faces with various patterns of transformations, including the variations in facial expressions, beards, skin colors, haircuts and poses.  This also illustrates the LGAN was able to disentangle different patterns of image transformations in its local coordinate charts on CelebA because of the orthonormality imposed on local tangent basis.

Moreover, we note that a face generated by LGAN could transform to the face of a different person in Figure~\ref{fig:celebA}. For example, in the first and the sixth row of the left figure, we can see that a female face transforms to a male face. Similarly, in the forth and the fifth row of the right figure, the male face gradually becomes more female.  This shows that local generators can not only manipulate attributes of input images, but are also able to extrapolate these inputs to generate very different output images.



\begin{figure}
\centering
\subfigure{
\begin{minipage}{1.0\linewidth}
\begin{center}
	   \includegraphics[width=0.95\linewidth]{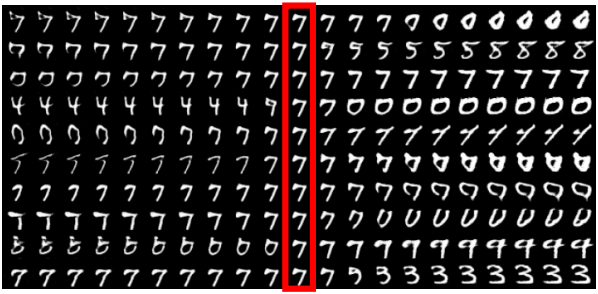}
\end{center}
\end{minipage}}
   \caption{Handwritten digits generated by LGAN on the MNITS dataset. The middle column in a red bounding box represents the image at the origin $\mathbf z=0$ of a local coordinate chart. In each row, the images are generated in one direction of a local coordinate.}\label{fig:mnist}
\end{figure}

We also illustrate the image generation results on the MNIST dataset in Figure~\ref{fig:mnist}. Again, we notice the factorized transformations in different tangent directions -- across different rows, the hand-written digits in the middle column changed to various writing styles. Also, a digit could gradually change to a different digit.  
This shows local coordinate charts for different digits were not isolated on the MNIST dataset.  Instead, they overlapped with each other to form a connected manifold covering different digits.


\subsection{Semi-Supervised Classification}

We report our classification results on the CIFAR-10 and SVHN (i.e., Street View House Number) datasets.

{\noindent\bf CIFAR-10 Dasetset.}
The dataset \cite{krizhevsky2009learning} contains $50,000$ training images and $10,000$ test images on ten image categories. We train the semi-supervised LGAN model in experiments, where $100$ and $400$ labeled examples  are labeled per class and the remaining examples are left unlabeled.
The experiment results on this dataset are reported by averaging over ten runs.

{\noindent\bf SVHN Dataset.} The dataset \cite{netzer2011reading} contains $32\times 32$ street view house numbers that are roughly centered in images.  The training set and the test set contain $73,257$ and $26,032$ house numbers, respectively. In an experiment, $50$ and $1,00$ labeled examples per digit are used to train the model, and the remaining unlabeled examples are used as auxiliary data to train the model in semi-supervised fashion.

\begin{figure}[t]
\centering
\subfigure{
\begin{minipage}{0.95\linewidth}
\begin{center}
	   \includegraphics[width=1.0\linewidth]{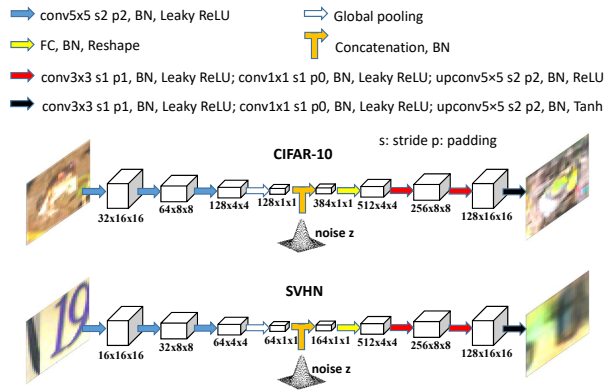}
\end{center}
\end{minipage}}
   \caption{Network architecture for local generators on SVHN and CIFAR-10.}\label{fig:arch}
\end{figure}

Figure \ref{fig:arch} illustrates the network architecture for local generators on both datsets.
For the discriminator, we used the networks used in literature \cite{salimans2016improved} to ensure fair comparisons on CIFAR-10 and SVHN datasets. In the appendix, we also present a larger convolutional network to train the discriminator that has been used in \cite{miyato2017virtual}, and we will show that the LGAN successfully beat the state-of-the-art semi-supervised models in literature  \cite{miyato2017virtual,laine2016temporal,park2017}.

In experiments, $100$ and $256$ local coordinates were used to train LGAN on SVHN and CIFAR-10, respectively, i.e., the noise $\mathbf z$ is a $100$-D and $256$-D vector. Here, more local coordinates were used on CIFAR-10 as natural scene images could contain more patterns of image transformations than street view house numbers.
To reduce computational cost, in each minibatch, ten coordinates were randomly chosen when computing the back-propagated errors on the orthonormal prior between local tangents. We also tested by sampling more coordinates but did not observe any significant improvement on the accuracy. So we only sampled ten coordinates in a minibatch iteration to make a balance between cost and performance.


\begin{table*}[]
\caption{Classification errors on both SVHN and CIFAR-10 datasets compared with the state-of-the-art methods. The error rates with $N_l=1000$ and $N_l=4000$ labeled training examples are reported. The best result is highlighted in bold. Note: *VAT did not report the deviation in the paper \cite{miyato2017virtual}. }
\label{tab:cifar10}
\begin{center}
\small\small
\begin{tabular}{c||c|c|c|c}    \toprule
\multirow{2}{*}{\emph{Methods}} & \multicolumn{2}{c}{SVHN}&\multicolumn{2}{|c}{CIFAR-10}\\\cline{2-5}
&\emph{$N_l=500$}& \emph{$N_l=1000$}&\emph{$N_l=1000$}& \emph{$N_l=4000$}   \\\midrule
Ladder Network \cite{rasmus2015semi} & --& --&  -- & 20.40$\pm$ 0.47\\
CatGAN \cite{springenberg2015unsupervised} &-- &-- & -- & 19.58$\pm$ 0.46 \\
ALI \cite{dumoulin2016adversarially} & -- & 7.41 $\pm$ 0.65 & 19.98 $\pm$ 0.89 & 17.99 $\pm$ 1.62\\
Improved GAN \cite{salimans2016improved} & 18.44 $\pm$ 4.8 & 8.11 $\pm$ 1.3 & 21.83 $\pm$ 2.01 & 18.63 $\pm$ 2.32 \\
Triple GAN \cite{li2017triple} & -- & 5.77$\pm$0.17 & -- & 16.99 $\pm$ 0.36\\
$\Pi$ model \cite{laine2016temporal} & 7.05$\pm$0.30 & 5.43$\pm$0.25 & -- & 16.55 $\pm$ 0.29\\
VAT \cite{miyato2017virtual}* & -- & 6.83 & -- & 14.87 \\
FM-GAN \cite{kumar2017improved} & 6.6$\pm$1.8& 5.9$\pm$1.4& 20.06 $\pm$ 1.6 & 16.78 $\pm$ 1.8 \\
LS-GAN \cite{qi2017loss} & --& 5.98 $\pm$ 0.27& -- & 17.30 $\pm$ 0.50\\\midrule
Our approach & \bf 5.48 $\pm$ 0.29 & \bf 4.73 $\pm$ 0.16 & \bf 17.44 $\pm$ 0.25 & \bf 14.23 $\pm$ 0.27 \\\bottomrule
\end{tabular}
\end{center}

\end{table*}

Table~\ref{tab:cifar10} reports the experiment results on both SVHN and CIFAR-10. On SVHN, we used $500$ and $1,000$ labeled images to train the semi-supervised LGAN, which is $50$ and $100$ labeled examples per class, and the remaining training examples were left unlabeled when they were used to train the model. Similarly, on CIFAR-10, we used $1,000$ and $4,000$ labeled examples with the remaining training examples being left unlabeled. The results show that on both datasets, the proposed semi-supervised LGAN outperforms the other compared GAN-based semi-supervised methods.

\begin{figure*}[htpb]
\centering
\subfigure[SVHN]{
\begin{minipage}{0.46\linewidth}
\begin{center}
	   \includegraphics[width=0.85\linewidth]{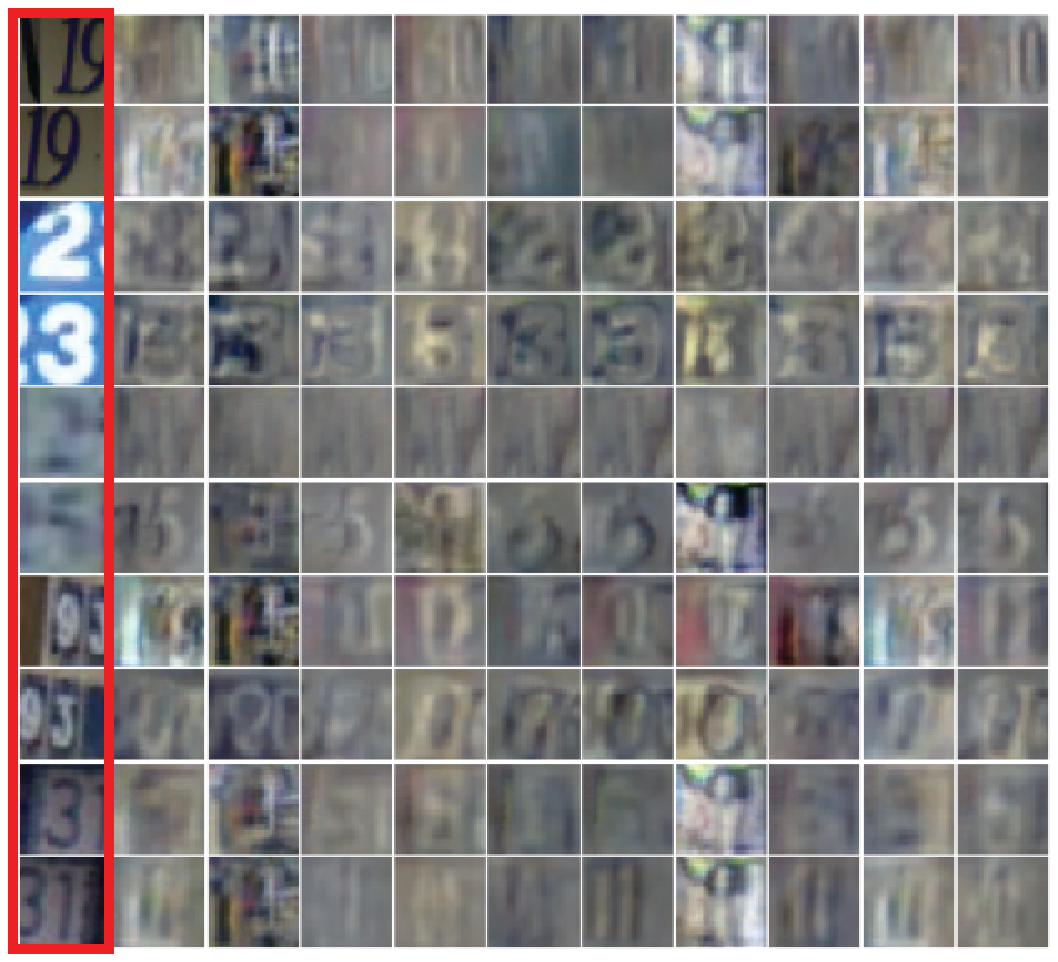}
\end{center}
\end{minipage}}
\subfigure[CIFAR-10]{
\begin{minipage}{0.46\linewidth}
\begin{center}
	   \includegraphics[width=0.85\linewidth]{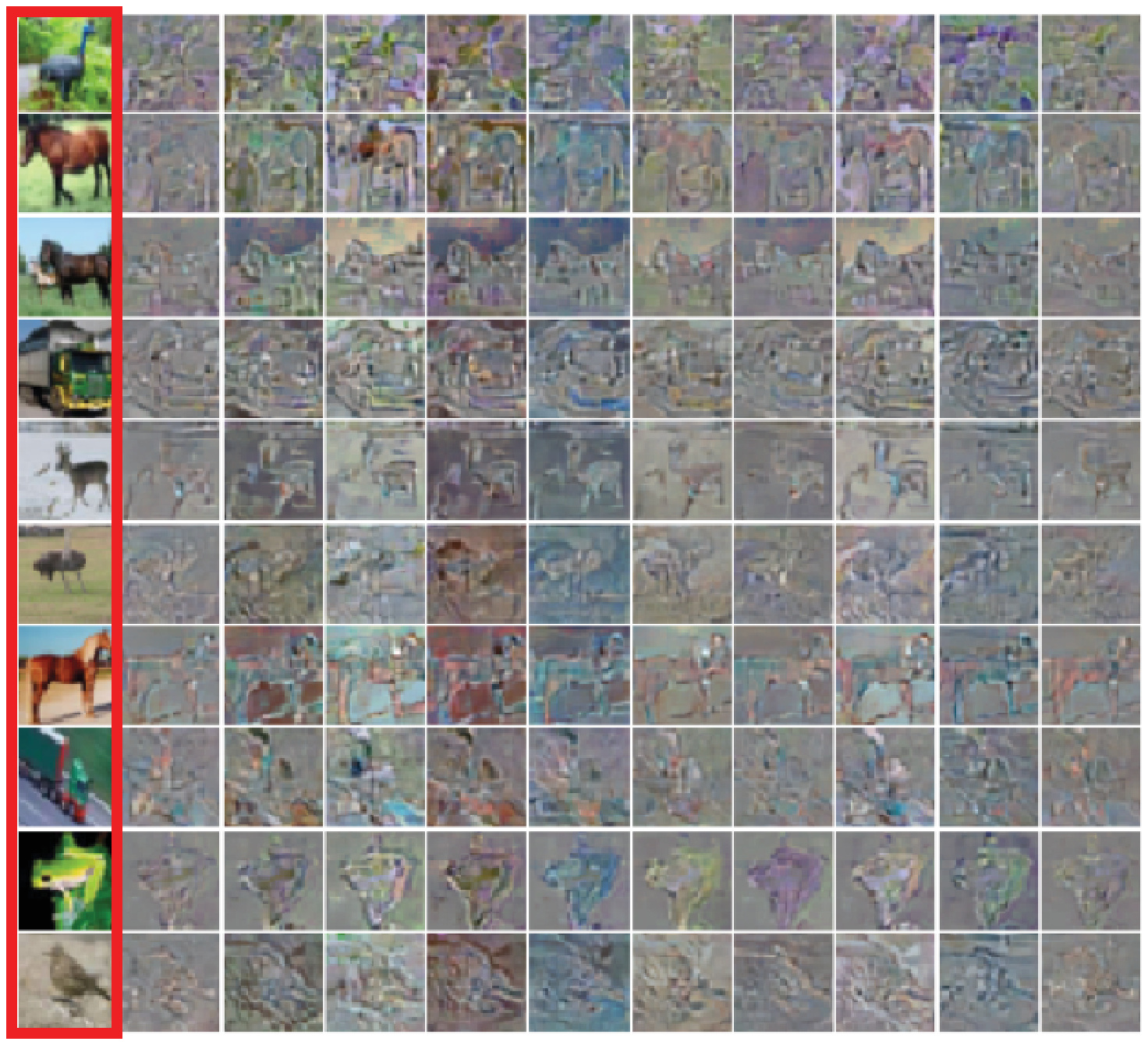}
\end{center}
\end{minipage}}
   \caption{Tangent images generated by LGAN along ten randomly chosen coordinates on SVHN and CIFAR-10 datasets. The first column in the red bounding box shows the original images, followed by their tangent images in each row. }\label{fig:tangents}
\label{Fig:dRNN}
\end{figure*}

Furthermore, we illustrate tangent images in Figure~\ref{fig:tangents} on SVHN and CIFAR-10 datasets.
The first column in the red bounding box shows the original images, followed by their tangent images generated by the learned local generators along ten randomly chosen coordinates in each row.
These tangent images visualize the local variations captured by LGAN along different coordinate directions.
This shows how the model is able to learn a locally consistent classifier by exploring the geometry of image transformations along these tangent directions in a neighborhood of the underlying manifold.

%

\section{Conclusion}
This paper presents a novel paradigm of localized GAN (LGAN) model along with its application in semi-supervised learning tasks. The model uses an atlas of local coordinate charts and associated local generators to cover an entire manifold, allowing it to capture distinct geometry of local transformations across the manifold. It also enables a direct access to manifold structures from local coordinates, tangents to Jacobian matrices without having to invert the global generator in the classic GAN. Moreover, by enforcing orthonormality between tangents, it can prevent the manifold from being locally collapsed to a dimensionally deficient subspace, which provides a geometric insight into alleviating mode collapse problem encountered in literature.  Its application to semi-supervised learning reveals the connection with Laplace-Beltrami operator on the manifold, yielding a locally consistent classifier resilient against perturbations in different tangent directions. Experiment results on both image generation and classification tasks demonstrate its superior performances to the other state-of-the-art models.

\appendix

\section{More Results on Semi-supervised Learning}

This section describes more experiment results on the semi-supervised learning using LGAN. Besides the small discriminator structure employed in Section 5.3, we further test LGAN with a larger CNN architecture, which is the same one as the ``Conv-Large" used in \cite{miyato2017virtual}. For convenience, we will refer this larger discriminator as Conv-Large, while the one used in Section~5.3 as ``Conv-Small" in the following. We compare the Conv-Large LGAN with the state-of-the-art semi-supervised learning methods (which are not necessarily the GAN-based) and report the results on both CIFAR-10 and CIFAR-100 datasets in this section. The architecture of the generator will keep the same as that used in Conv-Small experiments.

\subsection{Discriminator Architectures}

\begin{table}[b]

\begin{center}
\begin{tabular}{l l}
\hline
Name & Description \\
\hline\hline
Input & $32 \times 32$ RGB image \\
drop1 & Dropout $p=0.2$ \\
conv1a & 96, $3 \times 3$, pad=1, stride=1, LReLU \\
conv1b & 96, $3 \times 3$, pad=1, stride=1, LReLU \\
conv1c & 96, $3 \times 3$, pad=1, stride=2, LReLU \\
drop2 & Dropout $p=0.5$ \\
conv2a & 192, $3 \times 3$, pad=1, stride=1, LReLU \\
conv2b & 192, $3 \times 3$, pad=1, stride=1, LReLU \\
conv2c & 192, $3 \times 3$, pad=1, stride=2, LReLU \\
drop3 & Dropout $p=0.5$ \\
conv3a & 192, $3 \times 3$, pad=0, stride=1, LReLU \\
conv3b & 192, $1 \times 1$, LReLU \\
conv3c & 192, $1 \times 1$, LReLU \\
pool1 & Global mean pooling $6 \times 6 \rightarrow 1 \times 1$ \\
dense & Fully connected $192 \rightarrow 10$ \\
output & Softmax \\
\hline
\end{tabular}
\end{center}
\caption{The network architectures of Conv-Small}
\label{t:Conv-small}
\end{table}

\begin{table}[b]
\begin{center}
\begin{tabular}{l l}
\hline
Name & Description \\
\hline\hline
Input & $32 \times 32$ RGB image \\
drop1 & Dropout $p=0.2$ \\
conv1a & 128, $3 \times 3$, pad=1, stride=1, LReLU \\
conv1b & 128, $3 \times 3$, pad=1, stride=1, LReLU \\
conv1c & 128, $3 \times 3$, pad=1, stride=1, LReLU \\
pool1 & Maxpooling $2 \times 2$ \\
drop2 & Dropout $p=0.5$ \\
conv2a & 256, $3 \times 3$, pad=1, stride=1, LReLU \\
conv2b & 256, $3 \times 3$, pad=1, stride=1, LReLU \\
conv2c & 256, $3 \times 3$, pad=1, stride=1, LReLU \\
pool2 & Maxpooling $2 \times 2$ \\
drop3 & Dropout $p=0.5$ \\
conv3a & 512, $3 \times 3$, pad=0, stride=1, LReLU \\
conv3b & 256, $1 \times 1$, LReLU \\
conv3c & 128, $1 \times 1$, LReLU \\
pool3 & Global mean pooling $6 \times 6 \rightarrow 1 \times 1$ \\
drop4 & Dropout $p=0.1$ \\
dense & Fully connected $128 \rightarrow 10$ \\
output & Softmax \\
\hline
\end{tabular}
\end{center}
\caption{The network architectures of Conv-Large}
\label{t:Conv-large}
\end{table}

Table \ref{t:Conv-small} and \ref{t:Conv-large} summarize the architecture of Conv-Small and Conv-Large, respectively. 
We also apply the weight normalization \cite{weighted_norm} to all convolutional and dense layers in both architectures.

\subsection{Training Details for Conv-Large}

\begin{table*}[t]
\begin{center}
\begin{tabular}{c c c}
\hline
Method & CIFAR-10 & CIFAR-100 \\
\hline\hline
$\Pi$ model \cite{laine2016temporal} & $12.36 \pm 0.31$ & $39.19 \pm 0.36$\\
Temporal Ensembling \cite{laine2016temporal} & $12.16 \pm 0.24$ & $38.65 \pm 0.51$\\
Sajjadi et al. \cite{sajjadi2016} & $11.29 \pm 0.24$ & -\\
VAT \cite{miyato2017virtual} & $10.55$ & -\\
VadD \cite{park2017} & $11.32 \pm 0.11$ & -\\
\hline
LGAN (Conv-Large) & $\mathbf{9.77 \pm 0.13}$ & $\mathbf{35.52 \pm 0.33}$\\
\hline
\end{tabular}
\end{center}
\caption{Classification errors on both CIFAR-10 and CIFAR-100 with $4,000$ and $10,000$ labeled training examples respectively. The best result is highlighted in bold.}
\label{t:Results}
\end{table*}

Like in training the Conv-Small, we adopt Adam optimizer to train both the discriminator and generator. The learning rate is set to $4 \times 10^{-4}$, and the maximal training epoch is $1,200$. We gradually anneal the learning rates to zero during the last 400 epochs. The other settings are kept as same as those for training Conv-Small (Section~5.3). For CIFAR-100, which consists of $50,000$ $32 \times 32$ training images and $10,000$ test images in a hundred classes, we change the dropout rate of drop1 layer from $0.2$ to $0.1$ and the output dimension of the last layer to $100$.
We also adopt early stopping -- the training is terminated if the validation error stops decreasing over $100$ consecutive epochs after the $600$th epoch.
The two hyper-parameters $\mu$ and $\eta$ are chosen based on a separate validation set.


\subsection{Experimental Results for Conv-Large}

We compare the LGAN using Conv-Large discriminator with state-of-the-art semi-supervised baselines. The results are reported in Table \ref{t:Results}. Note that we used $4,000$ and $10,000$ labeled training examples for CIFAR-10 ($400$ images per class) and CIFAR-100 ($100$ images per class) respectively and the rest of training data unlabeled. From the table, we can see that LGAN with Conv-Large  outperforms the other compared methods on both datasets.

{\small
\bibliographystyle{ieee}
\bibliography{GAN,LSGAN,GANV1,egbib}
}

\end{document}